\documentclass[a4paper,fleqn]{cas-sc}

\usepackage[numbers]{natbib}

\begin{document}
\let\WriteBookmarks\relax
\def\floatpagepagefraction{1}
\def\textpagefraction{.001}
\shorttitle{Few-class Fidelity for CNN classifiers}
\shortauthors{W. Marchadour et~al.}

\title [mode = title]{Few-class Fidelity: Evaluating Explanations of Real-conditions CNN classifiers with Optimized Perturbations}                      
\tnotemark[1]

\tnotetext[1]{This work was partly funded by the ERA-Net CHIST-ERA grant [CHIST-ERA
19-XAI-007] long term challenges in ICT project INFORM (ID: 93603), and the Brittany Region.
}


\author[1,2]{Wistan Marchadour}[orcid=0000-0002-3450-9783]
\cormark[1]
\ead{wistanmarchadour@yahoo.fr}

\credit{Methodology Design, Software Development, Article Redaction}

\author[1]{Pedro Soto Vega}[orcid=0000-0001-5396-8531]

\credit{Models Training}

\author[2]{Franck Vermet}[orcid=0000-0003-3816-5401]
\ead{franck.vermet@univ-brest.fr}

\credit{Mathematical Validation, Article Review}

\author[1]{Mathieu Hatt}[orcid=0000-0002-8938-8667]
\cormark[2]
\ead{hatt@univ-brest.fr}

\credit{Methodology Design, Article Review}

\affiliation[1]{organization={LaTIM, INSERM, UMR 1101, Univ Brest},
                city={Brest},
                country={France}}

\affiliation[2]{organization={LMBA, CNRS, UMR 6205, Univ Brest},
                city={Brest},
                country={France}}

\cortext[cor1]{Corresponding author}
\cortext[cor2]{Principal corresponding author}


\begin{abstract}
The wide use of Convolutional Neural Networks (CNN) in numerous domains and real-world classification applications is justified by their high precision and automation speed, helping users concentrate on higher-expertise tasks. To better understand the models and avoid bias during deployment, eXplainable Artificial Intelligence (XAI) techniques can be used after training. But as the list of XAI solutions expand, comparisons between them diverge, and consensus over their evaluation cannot be reached. This paper proposes a variation of Fidelity-based XAI metrics, with a focus on real-conditions applications, where the number of classes is often low. The approach generates in-distribution, uncertainty-provoking perturbations, to ensure proper measurement of the XAI methods faithfulness. As demonstration of the evaluation framework's usefulness, it is compared with human-centric object localization and segmentation metrics. Once applied to both medical and natural imaging applications, it highlights the intricate correlation between domain, data curation, and XAI solution choices in order to validate training of a new CNN model.


\end{abstract}


\begin{highlights}
\item Identification of issues regarding Fidelity-based XAI evaluation for real-world, few-classes applications
\item Redefinition of the Fidelity metric concept, adapted to the observed constraints
\item Design of a novel framework for realistic perturbations generation
\item Application of the contributions to two different imaging domains
\end{highlights}


\maketitle

\section{Introduction}





The usefulness of Deep Learning (DL) for the automation of imaging tasks is now fairly demonstrated, with continuously improving performances. Latest publications often include some kind of interpretation of the DL models in order to increase transparency regarding their inner workings, which falls under the denomination eXplainable Artificial Intelligence (XAI). This topic is crucial for future deployment of DL-based methods, as without a clear explanation regarding how models reach their decisions, end-users may not have enough understanding and trust to integrate these outputs into real-world decision-support tools. Most of XAI algorithms, considered as "local interpretability", rely on analyzing individual input-output pairs to explain a specific outcome \cite{goyal2019counterfactual, petsiuk2018rise, ribeiro2016should, selvaraju2017grad}, by identifying the most salient pixels in the input image, with respect to the decision. Due to the large number of proposed methods using this concept, quantitative comparison is warranted, which led to several metrics being proposed, e.g., exploiting perturbations of the input image \cite{ancona2017towards, hooker2019benchmark, petsiuk2018rise} to assess the fidelity of the proposed explanations towards the classifier.

Although no definitive consensus has been reached so far to identify the best explainer against classifiers black-box behavior, several drawbacks have been observed regarding the metrics, such as the generation of Out-of-Distribution (OoD) perturbations \cite{qiu2021resisting, vsimic2025comprehensive}. In this paper, we propose a variation of current fidelity-based metrics, completed by an optimized perturbation generation framework. This metric leverages on the OoD issue, but also on a newly-defined constraint regarding few-class uncertainty, and is designed to evaluate XAI methods on real-conditions classification tasks. We demonstrate the usefulness of our framework on two different domains of application, with different imaging modalities and number of classes, and present the resulting rankings of a subset of currently available XAI methods.



\section{Related Works}

This study focuses on the evaluation of local, passive XAI methods, as defined in \cite{zhang2021survey}, relying on image perturbations and Fidelity measurement. Passive XAI algorithms are used only after the classifier is completely trained, making them applicable to any past, present and future DL models. Additionally, local XAI is used to analyze model reasoning "by the example", assigning each input parameter (i.e., pixels or voxels for 2D or 3D images) a value that weighs its relative importance with respect to the predicted classification. By combining both aspects, XAI methods yield attribution (or saliency) maps, that can be superposed to the input image for identification of its most relevant (regarding the classification output) parts. The local passive sub-domain of XAI has been highly explored \cite{fong2017interpretable, kapishnikov2019xrai, simonyan2013deep, smilkov2017smoothgrad, sundararajan2017axiomatic, zhang2021survey}, because these methods often require short computation time compared to other algorithms, and saliency maps are visually easier to comprehend.

The evaluation of the relevance of these saliency maps can be performed with respect to different contexts, that can be categorized into human-driven or human-independent. On the one hand, human-driven evaluation is either qualitative, with series of questions asked to a group of users, or quantitative, by comparing XAI maps with handcrafted annotations. In both cases, the alignment between explanations and human opinion is measured, which requires non-negligible resources (large enough human cohort and/or time to annotate data). Human-driven evaluation is thus quite costly to carry out, in addition to being more or less sensitive to subjectivity depending on the complexity and type of task. On the other hand, human-independent metrics are exclusively quantitative, and measure mathematical properties of XAI methods such as Fidelity \cite{fong2017interpretable, samek2016evaluating} and robustness \cite{lin2021you}, or validate some desired axioms \cite{sundararajan2017axiomatic}. This type of evaluation ensures that saliency maps are not manipulated to focus on human alignment performance, creating misplaced trust from external users, as proven before \cite{adebayo2018sanity}.
Because human-driven and human-independent evaluations are complementary, some studies, including ours, combine them to find XAI methods that are both clear and reliable \cite{chang2018explaining, fong2017interpretable, qiu2021resisting}.

\subsection{Perturbation-based Fidelity Metrics}

Fidelity evaluation frameworks, also described as Faithfulness metrics, are part of the human-independent metrics. They evaluate the relevance of the relative quantitative value assigned to pixels in a saliency map with respect to the classifier's decision \cite{yeh2019fidelity}. The most popular approach for measuring Fidelity consists in perturbing the input image: if the values of one or multiple pixels, deemed important by an XAI method, are altered "enough" (and what can be considered "enough" has to be determined), then the model's performance is excepted to drop, compared to using the image with the original values. On the contrary, altering the values of pixels evaluated as not relevant for the classification in the input image should not change the performance of the model, at least not significantly. Frameworks for image perturbation can be described across multiple axes:
\begin{itemize}
    \item \textbf{Which pixels should be altered?} Founding works on the subject selected random pixels to modify \cite{selvaraju2017grad}, observing the resulting output classification to infer on the usefulness of sampled features. Later, targeted solutions were designed, using the order of attribution values to find which pixels are most likely to induce a change in performance \cite{ancona2017towards, nam2020relative}. Most recent approaches use both ascending and descending attribution orders \cite{brocki2022evaluation}, with the drawback of an increased computation time.
    \item \textbf{How many pixels should be altered?} Ideally, images are altered one pixel at a time \cite{ancona2017towards}, resulting in more reliable metric evaluation. For low-resource systems, some alternatives occlude entire regions, either by patching the image \cite{montavon2018methods, samek2016evaluating}, or applying a segmentation step as a pre-processing, prior to the perturbation itself \cite{rieger2020irof}.
    \item \textbf{Which value should replace the original?} "Removing" the pixel information entirely (i.e., replacing by a "not a number" value) obviously cannot be used, therefore an alternative and less-informative value has to be set. Preliminary approaches used fixed replacement values such as e.g., black and white-corresponding values or the mean pixel intensity of the image \cite{ancona2017towards, samek2016evaluating}, until more adaptive solutions were designed, mostly due to the so-called Out-of-Distribution issue \cite{agarwal2020explaining, khamaiseh2022adversarial, qiu2021resisting}.
\end{itemize}

\subsection{Out-of-Distribution Issue}

\begin{figure}
    \centering
    \includegraphics[width=0.6\linewidth]{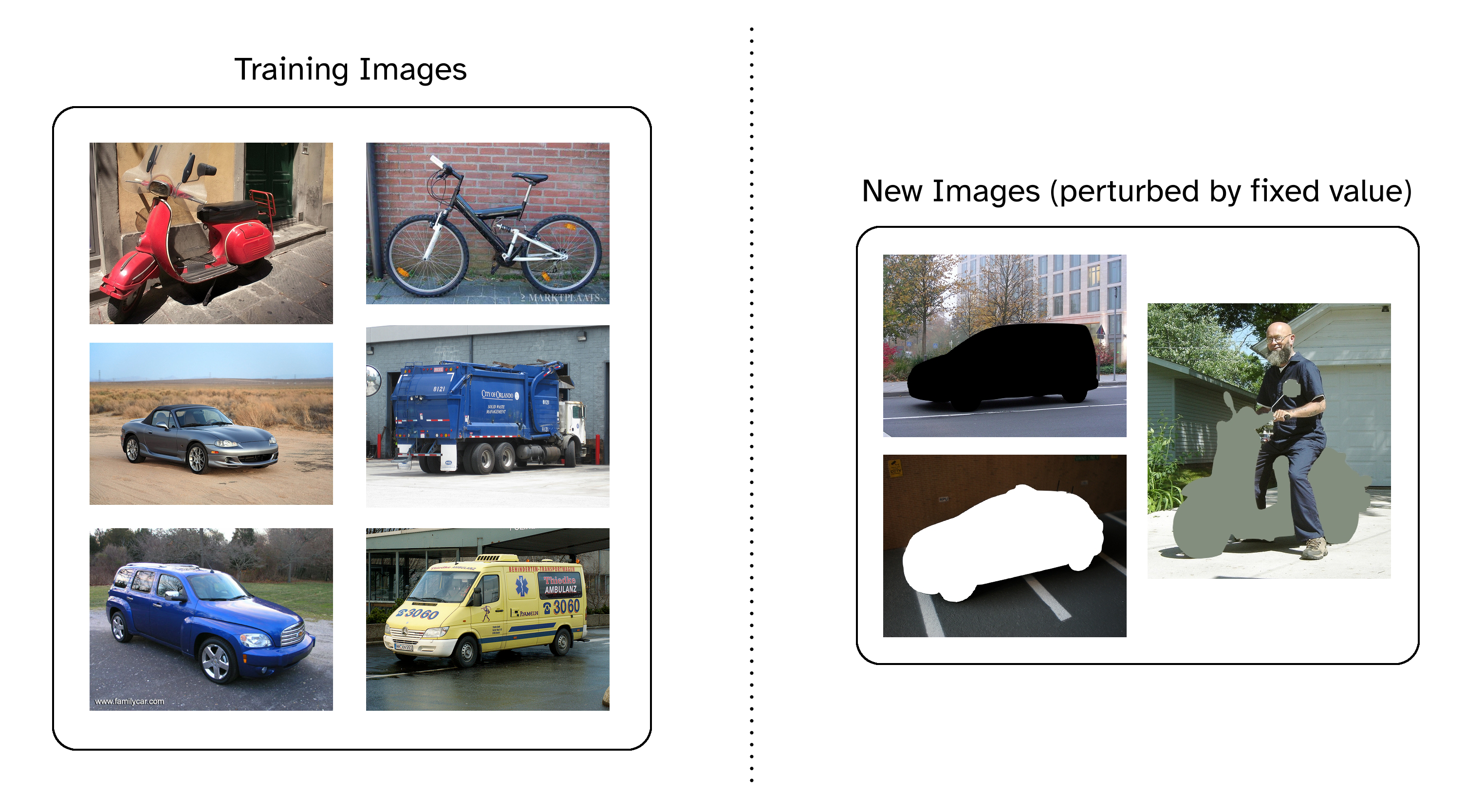}
    \caption{Illustration of Out-of-Distribution issue. When images altered using a fixed value are given as an input to the model, it cannot recognize any known class, which is not a desired effect for the purpose of fidelity estimation.}
    \label{fig:OoD_demo}
\end{figure}

In recent works, an important issue has been identified in some perturbation-based frameworks and denoted "Out-of-Distribution" (OoD) issue \cite{fong2017interpretable, hooker2018evaluating, nguyen2015deep}. It is directly associated with the choice of replacement values: When an input image is partly or entirely altered, by replacing its pixels values by a fixed, pre-defined value, the resulting image can be considered by the model as being entirely outside of the distribution of the samples used for its training (Figure \ref{fig:OoD_demo}). As a result, this OoD sample, when inferred on by the model, leads to an erreneous classification, not because important pixels have been changed (as intended), but rather because the model does not recognize the sample. This was determined as an easy and quick way to fool a deep network \cite{nguyen2015deep} and was then labeled as problematic for perturbation-based XAI methods \cite{fong2017interpretable} and other evaluation metrics relying on a similar concept\cite{hooker2018evaluating}.

Current solutions against OoD samples all rely on variations of dynamic pixels values replacement, while maintaining the objective of uninformative perturbation. Fong et al. \cite{fong2017interpretable} used Gaussian noise or blur to generate replacement values, thereby diminishing the level of information while keeping samples close to the training distribution. Agarwal et al. \cite{agarwal2020explaining} trained an "in-painting" generative model to replace objects of interest by their natural background in the image: while object information is correctly removed, it is not easily applicable in all contexts (especially in medical imaging), and implies the need to explain this new model alongside the classifier. Hooker et al. re-trained  the model to include OoD samples in the new dataset distribution \cite{hooker2019benchmark}, but this entails diverging from the original weights. Finally, closer to our approach, Qiu et al. \cite{qiu2021resisting} created an OoD detector, weighing samples according to their proximity with the training data distribution.



\section{Few-class Fidelity Metric}

\subsection{Few-class applications issue}

The Fidelity evaluation approach we present also addresses a new issue observed from real-world applications. In the literature, most studies are evaluated in the context of a large number of classes (orders of magnitude 10, 100, 1000), supported by very large image datasets. However, in specific domains, classification tasks often consider a limited number of classes, down to 2 (e.g. foreign object detection on industry lines \cite{zainab2023foreign} or glaucoma binary classification in medicine \cite{thanki2023deep}), without the need to further detail predictions with sub-categories. This difference, while barely noticeable during saliency maps computation, has a great impact on the perturbation of input images, especially with respect to the measurement of faithfulness.

In the first - ideal - situation above, perturbation of the classifier for XAI evaluation means changing the model output from the true class to any of the many others available, as it will lower the accuracy and be considered a "randomized" prediction. Now, among the few-class applications, let us take the binary classification case, which is often considered in medical imaging applications (e.g., benign vs. malignant lesion, advanced vs. early staging, no responder vs. responder...). Here, the expected output change actually corresponds to a "class switch" between the true class and the only other possible predicted class. Therefore, although an accuracy drop is indeed measured, the new prediction does not correspond to a random decision anymore. As detailed in other perturbation-based works \cite{petsiuk2018rise, qiu2021resisting, samek2016evaluating}, an input feature is relevant if its removal "reduces the prediction score of the classifier". We can thus conclude that perturbations should attempt to reduce the model's confidence (prediction scores), rather than making all its decisions faulty (accuracy).
Similarly, if we slightly increase the number of classes (up to 4-5), this switch issue is reduced by more prediction choices being available, but remains a concern for the reliability of accuracy-based faithfulness assessment.

\subsection{Uncertainty of the model for few-class applications}

This few-class issue described above, to our knowledge, has not been explicitly investigated in the context of Fidelity XAI evaluation. Some sections of proposed measurements could help resolving it, like the "random chance" probability by Agarwal et al. \cite{agarwal2020explaining}, but no current framework is able to completely address the constraint. This can be partly explained by the absence of a single clear objective for perturbation-based metrics, as existing solutions start from similar but non-identical hypotheses.

Arising from the idea of reducing the confidence of the classifier when making predictions on altered samples, we first define the concept of decision uncertainty: a model is considered uncertain for an image if it is unable to make a confident classification, whether correct or not, on this specific input. Derived from this, we consider the perfect uncertainty situation for which all classes have the same output probability, and use it to declare the uncertainty objective \(U_{obj}=\frac{1}{N}\) (\(N\) the number of classes). In the Fidelity metric detailed below, when all relevant information is removed from an image, we expect the model to output probability scores close to \(U_{obj}\) for each class. This target value, as the base idea of our framework, is later used to find appropriate perturbations to apply, and ultimately to measure the ability of XAI methods to capture only important features.

\subsection{Fidelity Metric Core}

\begin{figure}
    \centering
    \includegraphics[width=1.0\linewidth]{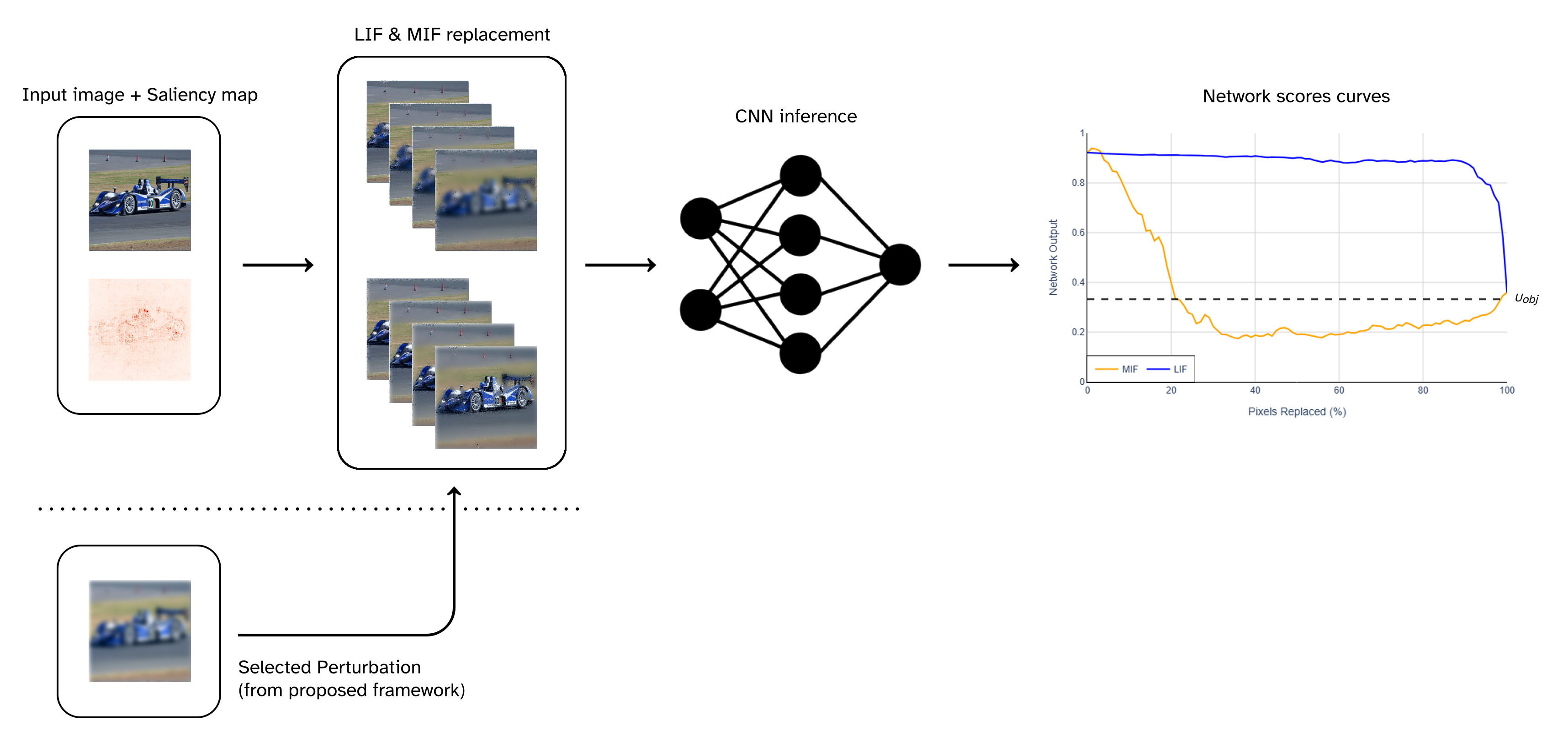}
    \caption{Proposed Fidelity metric. Input pixels are replaced iteratively over LIF and MIF orders (20\%/40\%/60\%/80\% shown here), altered images are inferred on by the classifier, and AUC of each order is measured. Finally, AUC values are compared to \(U_{obj}\) (dotted black line) to obtain \(Fid\).}
    \label{fig:Fidelity_process}
\end{figure}

Regarding the Fidelity metric itself, we based our approach on the iterative, dual-ordered replacement of input features, as presented in \cite{brocki2022evaluation, fong2017interpretable, petsiuk2018rise}, where output probability is monitored instead of accuracy. As illustrated in Figure \ref{fig:Fidelity_process}, the original input image is gradually altered before model inference, by replacing input features according to their attribution values in the corresponding saliency map. Both descending and ascending orders are used separately, respectively called Most and Least Important First (MIF and LIF). This allows not only important features are detected by the XAI methods, but also no relevant pixels are wrongfully missed, similarly to the F1-score measurement. After all steps of MIF and LIF orders are generated and inferred on, the output probabilities for one of the classes are monitored and plotted over the percentage of altered image, to measure the areas under these curves, named \(A_{MIF}\) and \(A_{LIF}\). In the reference works, \(A_{MIF}\) and \(A_{LIF}\) are simply subtracted to obtain the area between the curves as the Fidelity score of the tested saliency map, but the introduced \(U_{obj}\) target value requires changes in the formula:
\begin{itemize}
    \item For the LIF order, the close-to-perfect case implies maintaining an output probability of 1.0 for most of the replacement process, until the last (important) pixels change values and make the scores reach \(U_{obj}\): the area \(A_{LIF}\) should be as close as possible to 1.0.
    \item For the MIF order, output scores should drop to \(U_{obj}\) early during pixel replacement, and then remain at this value for the rest of the process: the ideal \(A_{MIF}\) is equal to \(U_{obj}\).
\end{itemize}

We add a class-specific division on the described operation, in order to have an updated Fidelity score constrained between 0 and 1, with higher values for the most faithful XAI methods:
\begin{equation} \label{eq:fidelity_final}
    Fid = 1 - \frac{\|1 - A_{LIF}\| + \|\frac{1}{N} - A_{MIF}\|}{1 + \frac{N-1}{N}}
\end{equation}
We chose to perform the replacements with a 1\% step, from 0\% (original image) to 100\% (all pixels altered). Considering that most datasets have much more than 100 pixels per image, it means that groups of pixels of various sizes, depending on the size of the input image, will be altered at each step, but without any a priori spatial neighboring correlation, unless segmentation being applied as a pre-processing step (see XRAI section below).

\subsection{Perturbation Framework}

As implied in the measurement of the Fidelity score, our process contains two fixed points, at 0\% and 100\% of image perturbation, with respective output scores of 1.0 and \(U_{obj}=\frac{1}{N}\). Exact values are not mandatory, as all saliency maps related to one input image will achieve the same scores on these fixed ends of MIF and LIF curves, enabling a fair comparison between them. While the 0\% point is inherently validated by inferring on the original image, the 100\% state requires to find an altered version of the image, that obtains approximately \(\frac{1}{N}\) probability for the investigated class, as well as for the other classes to validate the newly defined uncertainty constraint.

This suitable perturbation must also respect the OoD issue, by generating inputs that belong to the known distribution of the network. Although a direct value/distribution comparison is possible using Kullback-Leibler divergence \cite{anders2020fairwashing}, it becomes exponentially more expensive and unusable when dealing with entire image datasets. A simpler and faster way to detect potential OoD samples is to assess the similarity between the original image and its altered version, ensuring proximity to the data distribution as much as possible.

To summarize our objectives to address both OoD and few-class issues, we consider a perturbation as suitable for the original image if (1) it shows high structural similarity to this image, to be considered at least "near" the known dataset of the classifier, and (2) it is able to generate an uncertainty of the model, by forcing it to make no confident predictions once the image is fully altered. From current knowledge on automatic perturbation frameworks, it appears only image generation processes \cite{agarwal2020explaining, chang2018explaining} could potentially respect such strong constraints. However, their DL-based (thus black-box) structure would require additional explainability efforts, which does not match our goal for a fully-transparent solution. Instead, we propose to use a heuristic, non-DL-based framework for optimized alteration. After generating multiple alteration candidates for 100\%-replacement of a given image, we apply quantitative metrics to rank them according to our criteria above, and choose the most appropriate alteration for each tested image, for Fidelity assessment of XAI methods.

\subsubsection{Candidates Generation}

Proven by works on adversarial attacks for deep neural networks \cite{jiang2020poisoning, khamaiseh2022adversarial, moosavi2017universal}, it is known that one-shot, well-designed perturbations can add a random effect to images, and heavily decrease the performance of classifiers. However, in order to obtain a specific amount of confidence decrease as we seek with \(U_{obj}\), it rather requires a range of perturbation attempts, large enough to ensure at least one alteration candidate satisfies both OoD and few-class requirements.

From common non-DL image alteration techniques, we chose to apply Gaussian blurring to the original images, generating low-resolution versions that maintain the input structure, and are more similar than noise addition or single-value replacement. For some images, no suitable perturbation can be identified using blur only. In these cases, other alterations must be added to further increase the range of candidates, such as element-by-element mathematical operations, e.g., square value or product/division by mean value of the image, applied either before or after blurring. Finally, as we observed large differences in pixel values between original and modified images, each alteration combination aforementioned can have one of two additional overall adjustments: range or histogram matching between candidates and their source image.

\begin{figure}
    \centering
    \includegraphics[width=1.0\linewidth]{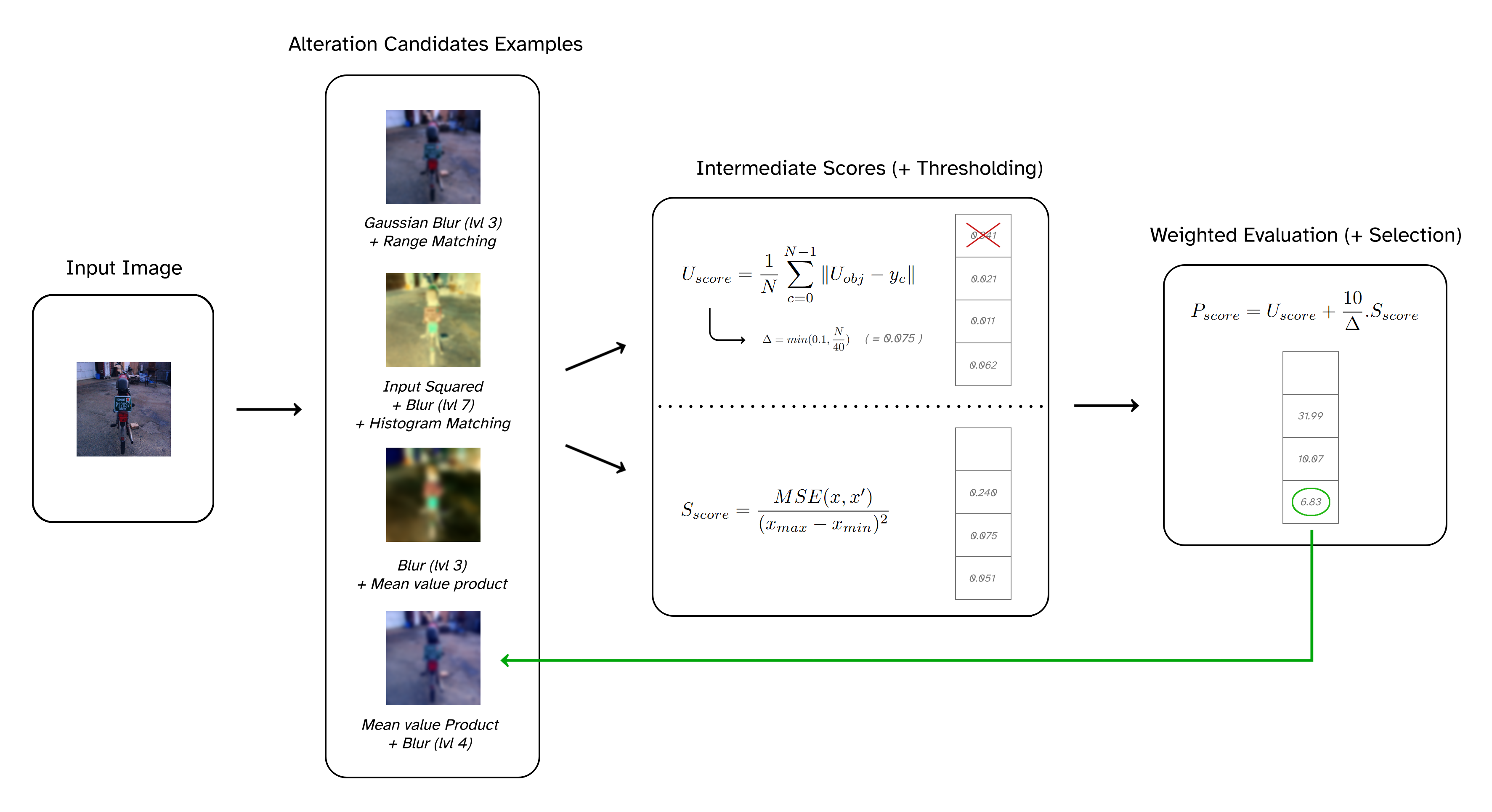}
    \caption{Process of perturbation candidate selection. Candidates are generated from combining various image modifications (four examples given). \(U_{score}\) is computed for all candidates, and scores above \(\Delta\) are discarded. \(S_{score}\), then \(P_{score}\) are applied on remaining possible alterations. Candidate with the lowest \(P_{score}\) is selected as perturbation for the Fidelity metric, for the corresponding input image.}
    \label{fig:Perturbation_framework}
\end{figure}

This grid of alterations leads to 33 possible variations, on which we apply a varying range of standard deviation values for the Gaussian blur (12 to 80 iterations depending on their influence on the final alteration result). In total, this process generates 804 alteration candidates for each tested image. Four of them are shown in Figure \ref{fig:Perturbation_framework}. From this list of potential perturbations, only one will be used in the Few-class Fidelity metric. This extended list, though it may be reduced for future versions, offers a very high probability to always find an optimized alteration that satisfies our required constraints.

\subsubsection{Perturbation Selection}

Because two different requirements were presented (based on few-class and OoD issues), separate measurements are required to validate them quantitatively. Therefore, the final selected target perturbation will be the candidate that offers the best trade-off between these two weighted measures.

The ability of a candidate to generate uncertainty in the classifier's output is assessed by one inference pass of the model, then computing the average difference between each class score and \(U_{obj}\). Here, accuracy cannot be used (according to our design choices for the Fidelity metric), because it requires multiple images for its measurement, yet each candidate must be evaluated individually. This intermediate measure is defined as the Uncertainty score \(U_{score}\) (Figure \ref{fig:Perturbation_framework}):
\begin{equation} \label{eq:uncertainty_score}
    U_{score} = \frac{1}{N} \sum_{c=0}^{N-1} \|U_{obj} - y_{c}\|
\end{equation}
To ensure that the selected candidates achieve model uncertainty across all classes, it is possible to eliminate some alterations before the rest of the framework is applied, by defining a threshold \(\Delta\) over \(U_{score}\). If the value exceeds the threshold, it means the classifier still makes confident predictions, and the evaluated candidate should be discarded regardless of its future performance. We propose to set \(\Delta\) depending on the number of classes, with a maximum possible value:
\begin{equation} \label{eq:delta}
    \Delta = min(0.1, \frac{N}{40})
\end{equation}
This formula minimizes \(\Delta\) for N \(\epsilon\) [2; 3] because it is easier to achieve class output convergence towards \(U_{obj}\) when N is lower, but maintains a strong criteria for N \(\geq\) 4 to keep only acceptable \(U_{score}\) results. This pre-selection also optimizes an already large framework, by avoiding unnecessary computation times.

For the OoD issue assessment, we simplify candidates evaluation by comparing them with their source image, to obtain close, yet faster results than similarity with the entire dataset. The Mean Squared Error (MSE) is used to perform this task, as it is more sensitive to large pixel-value differences than other measures. We then define a Similarity score \(S_{score}\), equal to the MSE between an alteration candidate \(x'\) and its original image \(x\), divided by the range of \(x\) pixel values (Figure \ref{fig:Perturbation_framework}):
\begin{equation} \label{eq:similarity_score}
    S_{score} = \frac {MSE(x, x')} {(x_{max} - x_{min})^2}
\end{equation}
Division by the squared range enables consistent measurement of \(S_{score}\) across datasets and applications, including when images themselves do not have a finite range of pixel values, which is important for the weighted combination below. Lower \(S_{score}\) values indicate better pixel-to-pixel similarity, therefore candidates are more likely to be in-distribution for the classifier.

Once \(U_{score}\) and \(S_{score}\) are computed for each candidate, we combine the values to obtain the Perturbation score \(P_{score}\) (Figure \ref{fig:Perturbation_framework}): 
\begin{equation} \label{eq:perturbation_score}
    P_{score} = U_{score}+\frac {10} {\Delta}.S_{score}
\end{equation}
\(S_{score}\) is deliberately given a higher importance than \(U_{score}\) in this weighted evaluation, because the remaining candidates at \(P_{score}\) calculation already satisfy the uncertainty constraint (\(U_{score} < \Delta\)). The weight of \(S_{score}\) also depends on \(\Delta\) to match the orders of magnitude of \(U_{score}\) and \(S_{score}\) lowest values, and guarantees they both contribute towards \(P_{score}\) (any high score is not important as it will probably leads to eliminate the candidate). This formula was designed heuristically from experimenting on different datasets, and should be applicable to a large field of image types or classifier models without modifications. After every candidate is evaluated, only the one with the lowest \(P_{score}\) is kept, and paired with its source image for assessing the Fidelity of all saliency maps constructed from the original input (Figure \ref{fig:Fidelity_process} bottom).




\section{Experiments}

In the following experiments, we exploit our newly-defined Fidelity-based XAI metric by (1) providing the hierarchy over several saliency-based XAI methods for two applications with different imaging modalities, model architectures and number of classes, and (2) compare the rankings with a human-driven evaluation, independent of the presented framework.

\subsection{Setup}

\subsubsection{Dataset and Models}

\paragraph{ImageNet Vehicles}

We extracted a subset of ImageNet, containing only vehicle images, divided into three classes to stay in our few-class context: “private”, “two-wheeled”, and “trucks and services”. This custom dataset is fairly balanced, and contains a total of 20 800 images\footnote{The dataset is available at \url{https://www.kaggle.com/datasets/wistanmarchadour/imagenet-3classes}}. Two different DL architectures were trained using a standard train/validation/test split, based on ResNet-18 and DenseNet-121, reaching accuracy of 93.5\% and 92.6\% respectively. From the test group, we extracted 2500 images to use for XAI experiments, all correctly predicted by both models, to better compare models and avoid misdirected explanations.

\paragraph{Computed Tomography (CT) Contrast Agent Injection}

The second dataset combines the LIDC-IDRI \cite{armato2011lung} and LNDb \cite{pedrosa2019lndb} medical imaging public datasets, containing 1312 chest 3D CT-scans. We extracted 2072 2D slices, and used them to train two different architectures, namely ResNet-50 and Xception for detecting if contrast agent was used for the acquisition. This task is considered a "toy example" because it can be performed by even non-trained individuals, thus it offers an easy context for testing our Fidelity metric. Trained models reached accuracy of 99.1\% and 99.3\% for ResNet-50 and Xception, respectively. Regarding the XAI test group, it was created similarly as for the vehicles application, and is composed of 259 CT 2D slices.

\subsubsection{XAI Methods}

We performed the Fidelity evaluation and ranking on a selection of well-known and popular saliency-based XAI methods. They are mostly inspired on the gradients backpropagation concept, widely exploited for generation of local attribution maps. We deliberately left perturbation-based algorithms aside here, to avoid any metrics advantage (also designed on perturbation), though further studies could expand the list of tested techniques with this category. Among the available methods, we refined the list using estimated computation costs, and previous conclusions on potential biases \cite{adebayo2018sanity}, leading to the following choices: Backpropagation (BP) \cite{simonyan2013deep}, Deconvolution (Deconv) \cite{zeiler2013visualizing}, Expected Gradients (EG) \cite{erion2021improving}, Integrated Gradients (IG) \cite{sundararajan2017axiomatic}, and GradCAM \cite{selvaraju2017grad}. Details are required on some methods: in the case of IG, because the chosen baseline has a strong impact on the saliency maps, we used either the minimum, maximum, or zero value of the image as reference, creating three different variants (or less if some references are identical), called IG(min), IG(max) and IG(zero). Similarly, for EG computation, we randomly extracted 200 training images from the corresponding datasets, to obtain the required so-called "background set".

To the established list of XAI methods and variants, we add saliency maps alteration techniques that have demonstrated promising results, and may provide better metric scores: XRAI segmentation \cite{kapishnikov2019xrai} and SmoothGrad (SG) noise reduction (along with its unpublished Squared version SG²) \cite{smilkov2017smoothgrad}. For each parent method in the original list, one of these alterations can be added after saliency map generation, yielding 3 new children combinations. We note that for SG and SG², the noise level parameter was preemptively optimized (as required by the authors), using a different, smaller subset of test images. It was then found that a low value, between 10\% and 20\% of the image range of values, is the best option for standard deviation of the Gaussian noise.

\subsection{Fidelity of XAI methods}

The first experiment is the direct application of the metric on the presented datasets and XAI methods. On each XAI test group, the Perturbation framework is applied to select appropriate image alterations. Simultaneously, saliency maps are computed, then passed through the metric for quantitative evaluation.

\begin{table}
    \centering
    \begin{tabular}{>{\raggedright\arraybackslash}p{0.15\linewidth}|>{\raggedright\arraybackslash}p{0.15\linewidth}|>{\centering\arraybackslash}p{0.1\linewidth}|>{\centering\arraybackslash}p{0.1\linewidth}>{\centering\arraybackslash}p{0.1\linewidth}>{\centering\arraybackslash}p{0.1\linewidth}} \hline
         Model&  Method&  No alter.&  +XRAI&  +SG&  +SG²\\ \hline
         &  BP&  0.614&  0.603&  0.725&  0.713\\
         &  Deconv&  0.635&  0.632&  0.658&  0.633\\
         &  EG&  0.714&  0.697&  \underline{0.737}&  0.727\\
         ResNet-18&  IG(min)&  0.635&  0.650&  0.667&  0.669\\
         &  IG(zero)&  0.698&  0.684&  0.723&  0.720\\
         &  IG(max)&  0.686&  0.691&  0.706&  0.699\\
         &  GradCAM&  0.683&  0.688&  0.676&  0.676\\
         &  Random&  0.595&  &  &  \\ \hline
 & BP
& 0.611& 0.592& 0.720& 0.704\\
 & Deconv
& 0.680& 0.716& 0.679& 0.677\\
 & EG
& 0.727& 0.703& \textbf{0.746}& \textbf{0.743}\\
 DenseNet-121& IG(min)
& 0.638& 0.656& 0.678& 0.682\\
 & IG(zero)
& 0.703& 0.677& 0.719& 0.720\\
 & IG(max)
& 0.685& 0.686& 0.711& 0.700\\
 & GradCAM
& 0.675& 0.677& 0.684& 0.685\\
 & Random& 0.596& & & \\ \hline
    \end{tabular}
    \caption{Few-class Fidelity scores for ImageNet vehicles application.}
    \label{tab:Fidelity_vehicles}
\end{table}

\begin{figure}
    \centering
    \includegraphics[width=1.0\linewidth]{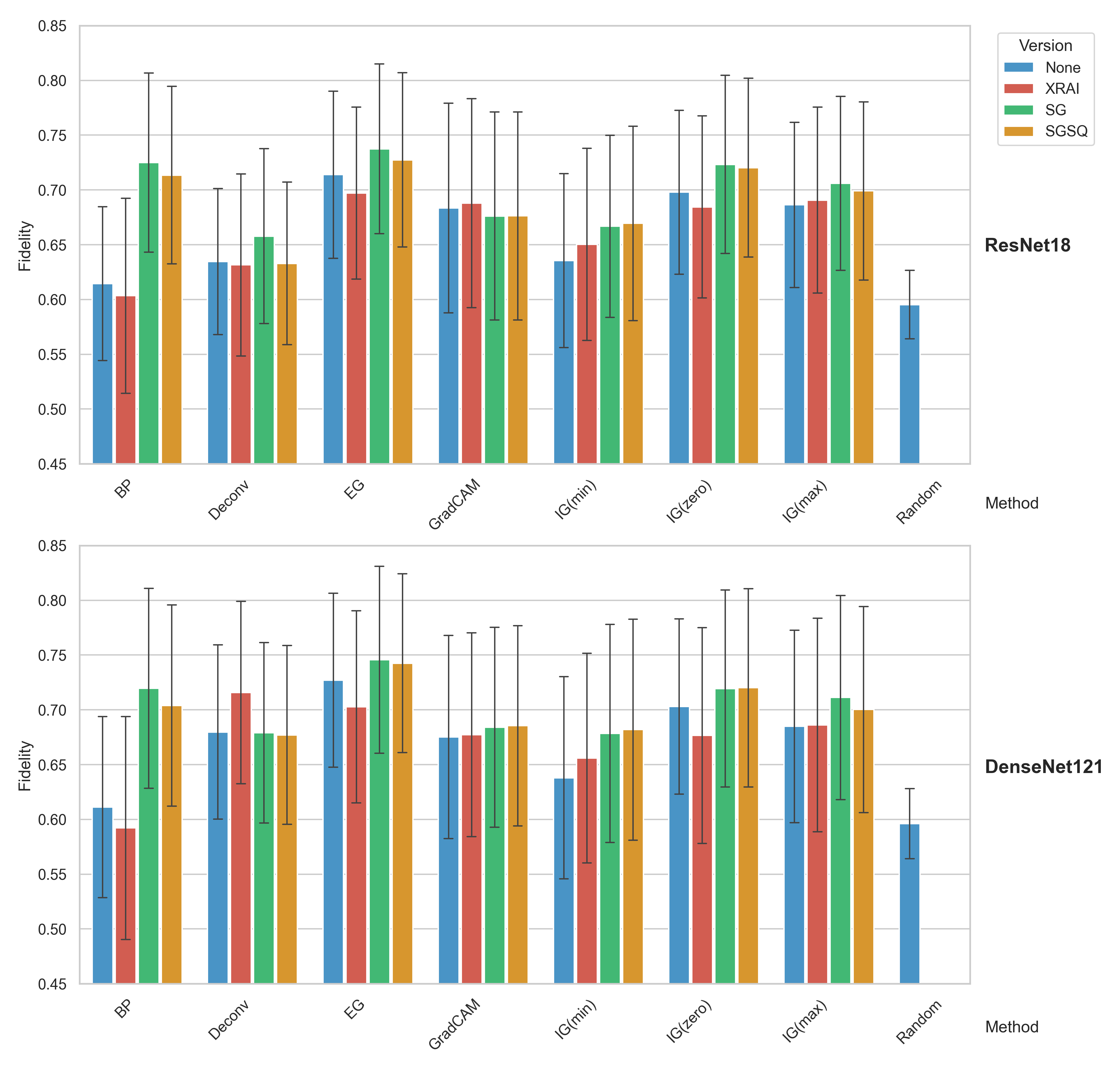}
    \caption{Few-class Fidelity box plot for ImageNet vehicles application, per model and base XAI method.}
    \label{fig:Fidelity_vehicles}
\end{figure}

For the ImageNet three-class vehicles classification, the noise reduction performed by SG (green) and SG² (orange) provide a small improvement of the metric scores, for most methods (Figure \ref{fig:Fidelity_vehicles}). On both trained models, EG achieved the best performance when combined to these saliency maps alterations, while BP offers results very similar to those of Random saliency maps.

\begin{table}
    \centering
    \begin{tabular}{>{\raggedright\arraybackslash}p{0.15\linewidth}|>{\raggedright\arraybackslash}p{0.15\linewidth}|>{\centering\arraybackslash}p{0.1\linewidth}|>{\centering\arraybackslash}p{0.1\linewidth}>{\centering\arraybackslash}p{0.1\linewidth}>{\centering\arraybackslash}p{0.1\linewidth}} \hline
         Model&  Method&  No alter.&  +XRAI&  +SG&  +SG²\\ \hline
         &  BP&  0.827&  \underline{0.918}&  0.833&  0.904\\
         &  Deconv&  0.668&  0.771&  &  \\
         &  EG&  0.817&  0.907&  0.813&  0.891\\
         ResNet-50&  IG(min)&  0.875&  \underline{0.918}&  0.846&  0.915\\
         &  IG(max)&  0.646&  0.728&  &  \\
         &  GradCAM&  \underline{0.918}&  0.871&  0.843&  0.846\\
         &  Random&  0.615&  &  &  \\ \hline
 & BP
& 0.764& \textbf{0.930}& 0.752& 0.905\\
 & Deconv
& 0.679& 0.836& & \\
 & EG
& 0.777& 0.911& 0.781& 0.891\\
 Xception& IG(min)
& 0.837& 0.923& 0.782& 0.876\\
 & IG(max)
& 0.665& 0.786& & \\
 & GradCAM
& \textbf{0.931}& 0.914& 0.911& 0.911\\
 & Random& 0.551& & & \\ \hline
    \end{tabular}
    \caption{Few-class Fidelity scores for Contrast agent application. SG and SG² were not applied to Deconv and IG(max) for performance and computation time reasons.}
    \label{tab:Fidelity_CT}
\end{table}

\begin{figure}
    \centering
    \includegraphics[width=1.0\linewidth]{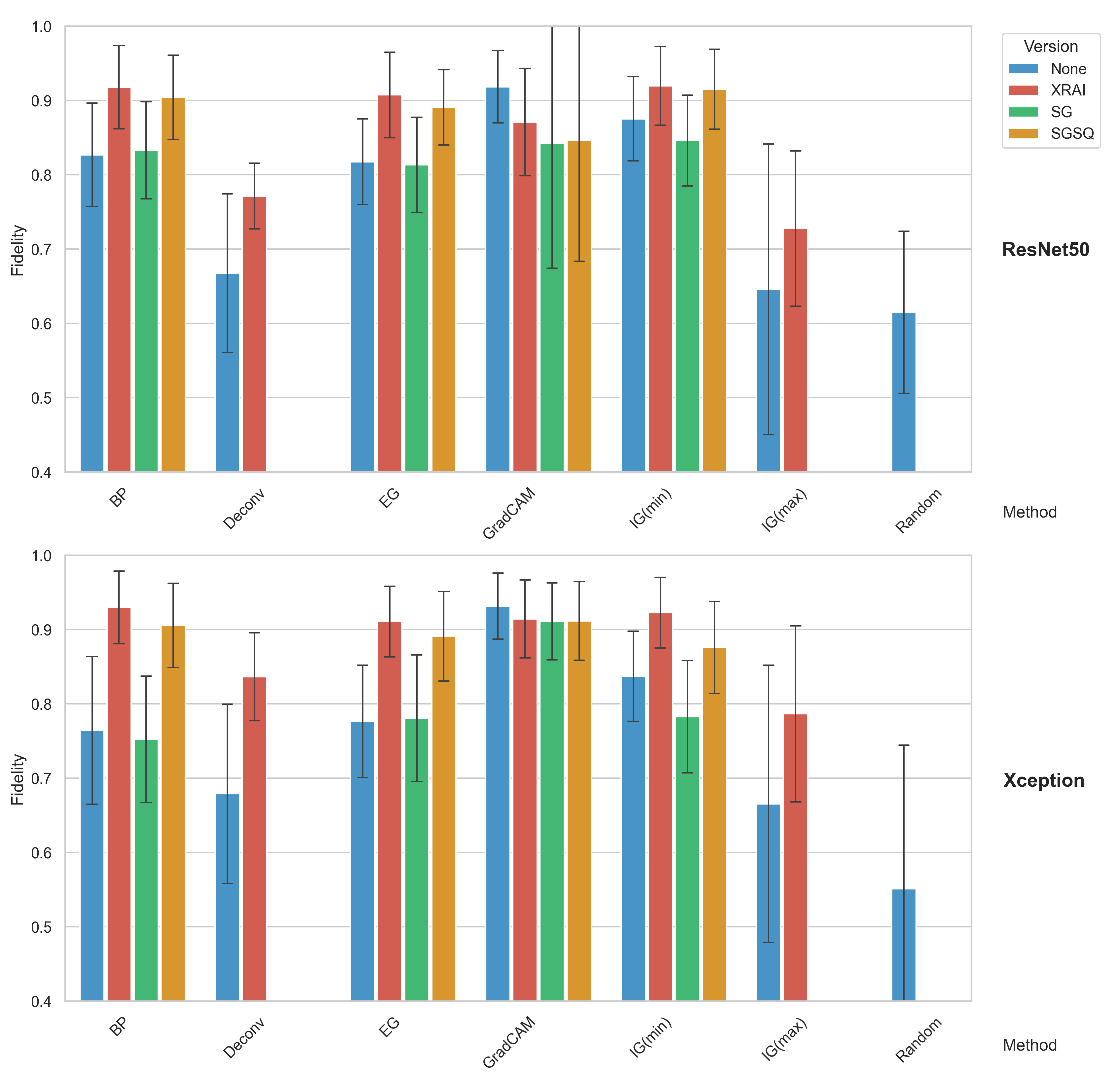}
    \caption{Few-class Fidelity box plot for Contrast agent application, per model and base XAI method. SG and SG² were not applied to Deconv and IG(max) for performance and computation time reasons.}
    \label{fig:Fidelity_CT}
\end{figure}

Regarding the binary Contrast Agent task, we observed a convergence between the networks in Figure \ref{fig:Fidelity_CT} results. However, on this dataset the highest Fidelity scores were often obtained with the XRAI segmentation algorithm (red), especially on BP and IG(min), on par with the vanilla GradCAM. Apart from the highest-ranking methods, other irregularities were found in the XAI hierarchy across applications: BP was in either higher or lower tier, and alterations had a fluctuating effect on saliency maps improvement.

\subsection{Object Localization / Segmentation}

To integrate human knowledge in our XAI evaluation study, and compare it to Fidelity results, we perform additional assessment on the ability of saliency maps to localize the most important input features. However, two separate protocols must be designed from the ground-truth of our datasets: while ImageNet vehicles annotations contain the bounding box of only one object (or the main object in the image), for the Contrast Agent task our truth consists of multiple non-rectangular regions that cannot be assembled.

\begin{table}
    \centering
    \begin{tabular}{>{\raggedright\arraybackslash}p{0.15\linewidth}|>{\raggedright\arraybackslash}p{0.15\linewidth}|>{\centering\arraybackslash}p{0.1\linewidth}|>{\centering\arraybackslash}p{0.1\linewidth}>{\centering\arraybackslash}p{0.1\linewidth}>{\centering\arraybackslash}p{0.1\linewidth}} \hline
         Model&  Method&  No alter.&  +XRAI&  +SG&  +SG²\\ \hline
         &  BP&  0.492&  0.526&  0.606&  0.594\\
         &  Deconv&  0.447&  0.505&  0.491&  0.481\\
         &  EG&  0.576&  \textbf{0.609}&  0.605&  0.599\\
         ResNet-18&  IG(min)&  0.528&  0.553&  0.561&  0.574\\
         &  IG(zero)&  0.562&  0.604&  0.600&  0.604\\
         &  IG(max)&  0.571&  0.603&  0.598&  0.597\\
         &  GradCAM&  0.577&  0.578&  0.587&  0.587\\
         &  Random&  0.324&  &  &  \\ \hline
 & BP
& 0.499& 0.531& 0.588& 0.582\\
 & Deconv
& 0.599& \textbf{0.634}& 0.612& 0.608\\
 & EG
& 0.591& 0.625& 0.617& 0.613\\
 DenseNet-121& IG(min)
& 0.540& 0.567& 0.575& 0.582\\
 & IG(zero)
& 0.564& 0.605& 0.586& 0.601\\
 & IG(max)
& 0.579& 0.604& 0.601& 0.585\\
 & GradCAM
& 0.576& 0.576& 0.575& 0.577\\
 & Random& 0.326& & & \\ \hline
    \end{tabular}
    \caption{Object localization scores for ImageNet vehicles application.}
    \label{tab:ObjLoc_vehicles}
\end{table}

\begin{figure}
    \centering
    \includegraphics[width=1.0\linewidth]{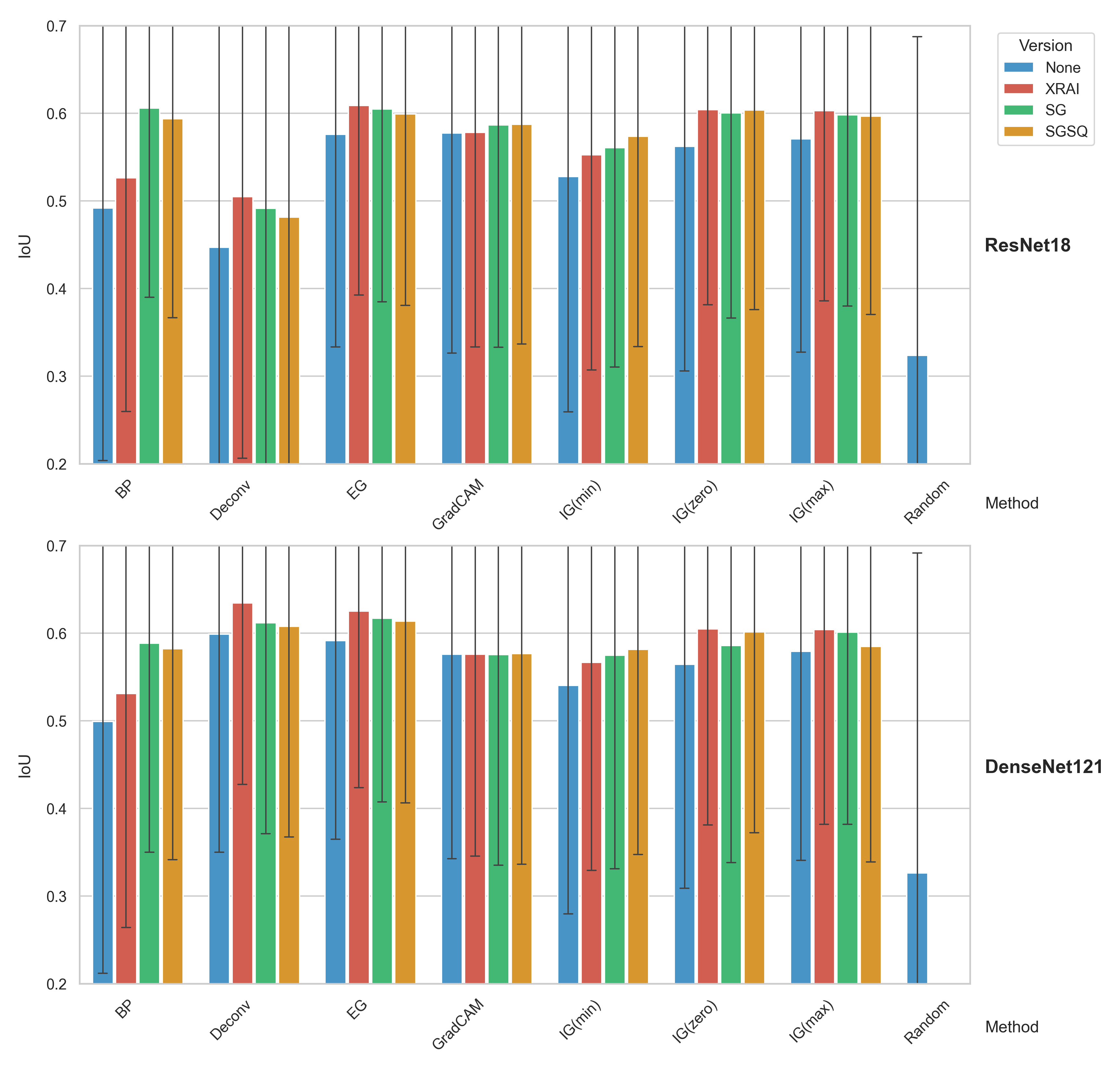}
    \caption{Object localization box plot for ImageNet vehicles application, per model and base XAI method.}
    \label{fig:ObjLoc_vehicles}
\end{figure}

\paragraph{ImageNet Vehicles}
The object localization experiment is inspired by the corresponding evaluation in \cite{selvaraju2017grad}. Saliency maps are binarized based on the area taken by the bounding box annotation, a box around the largest continuous region is drawn, and the Intersection over Union (IoU) is computed between truth and map boxes. Area-based binarization is modified from fixed-intensity threshold in the original version, to account for the varying size of objects in ImageNet images. Measured IoU scores distributions are similar across methods (Figure \ref{fig:ObjLoc_vehicles}), but remain better than fully random attributions. Vanilla average scores are almost always improved by maps alterations, though XRAI (red) is overall the best addition, consisten with using the largest region in IoU computation. The only method not benefiting from alterations is GradCAM, being a localization-based method already. Trained models have a convergence in hierarchy, and agree on EG+XRAI as most successful method, while Deconv+XRAI gets high IoU score only with DenseNet-121.

\begin{table}
    \centering
    \begin{tabular}{>{\raggedright\arraybackslash}p{0.15\linewidth}|>{\raggedright\arraybackslash}p{0.15\linewidth}|>{\centering\arraybackslash}p{0.1\linewidth}|>{\centering\arraybackslash}p{0.1\linewidth}>{\centering\arraybackslash}p{0.1\linewidth}>{\centering\arraybackslash}p{0.1\linewidth}} \hline
         Model&  Method&  No alter.&  +XRAI&  +SG&  +SG²\\ \hline
         &  BP&  0.453&  0.562&  0.476&  0.579\\
         &  Deconv&  0.083&  0.136&  &  \\
         &  EG&  0.317&  0.421&  0.368&  0.538\\
         ResNet-50&  IG(min)&  0.510&  0.589&  0.447&  \textbf{0.615}\\
         &  IG(max)&  0.085&  0.127&  &  \\
         &  GradCAM&  0.377&  0.386&  0.375&  0.374\\
         &  Random&  0.075&  &  &  \\ \hline
 & BP
& 0.307& \textbf{0.548}& 0.230& 0.460\\
 & Deconv
& 0.099& 0.158& & \\
 & EG
& 0.223& 0.333& 0.217& 0.338\\
 Xception& IG(min)
& 0.361& \textbf{0.552}& 0.238& 0.460\\
 & IG(max)
& 0.122& 0.176& & \\
 & GradCAM
& 0.379& 0.385& 0.320& 0.326\\
 & Random& 0.074& & & \\  \hline
    \end{tabular}
    \caption{Dice scores for Contrast agent application. SG and SG² were not applied to Deconv and IG(max) for performance and computation time reasons.}
    \label{tab:DSC_CT}
\end{table}

\begin{figure}
    \centering
    \includegraphics[width=1\linewidth]{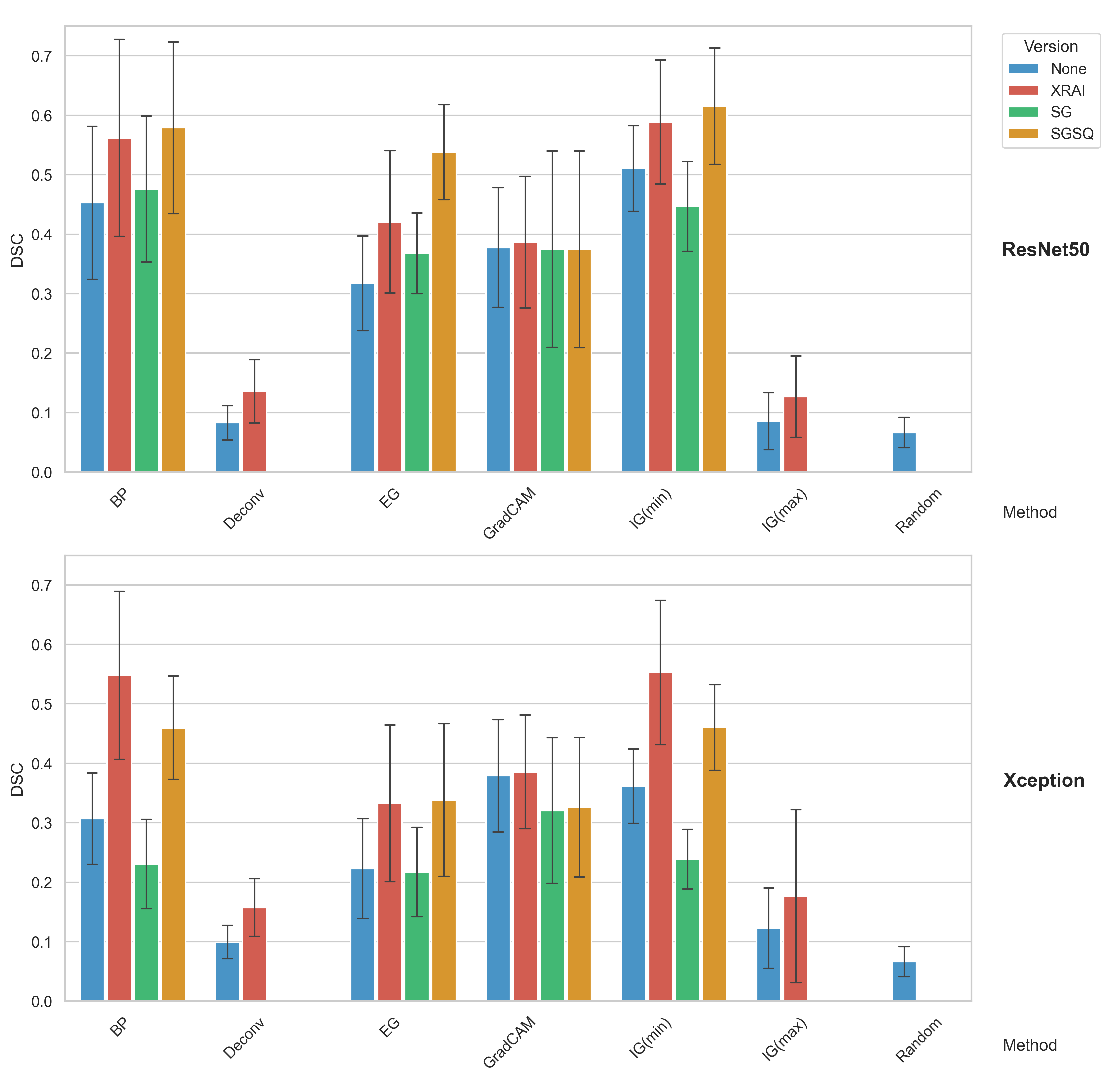}
    \caption{Dice bar plot for Contrast agent application, per model and base XAI method. SG and SG² were not applied to Deconv and IG(max) for performance and computation time reasons.}
    \label{fig:DSC_CT}
\end{figure}

\paragraph{Contrast Agent}
Given the multi-region ground-truth and lack of public annotations for this task, we manually segmented the most relevant regions of the image for human classification according to a clinical expert, i.e. the mediastinum area. Therefore, we preferred the Dice Similarity Coefficient (DSC), used in segmentation tasks evaluation, to measure human-driven performance of XAI methods, while preserving a similar process to the localization experiment above. Saliency maps are iteratively thresholded (1\% step), DSC is computed between percentage-specific binarized map and ground-truth mask, and the max DSC for each test image is retained before average. From results in Figure \ref{fig:DSC_CT}, we first note close-to-random performance from Deconv and IG(max), emphasizing the importance of reference value choice for the IG method. Then, BP+XRAI and IG(min)+XRAI offer the most faithful saliency maps, completed by IG(min)+SG² for ResNet-50 model. On the other hand, the SG alteration did not prove useful.



\section{Discussion}

The experiments of this study were designed to assess the performance of a wide range of XAI methods, while varying the application, model, image type, and metric used.

From the Fidelity results alone, we find high-end scores belonging to the same sub-list of methods. Therefore, it is clear the relevance of XAI methods is influenced by the studied application (and its corresponding dataset), not by the models architecture. This observation is consistent with literature statements, showing that dataset characteristics are key drivers of CNN performance, rather than the actual network components \cite{shin2016deep}.

Now, when comparing object localization / segmentation experiment results with their Fidelity counterpart, a strong correlation appears in model-wise XAI methods rankings, and for the impact of alteration techniques. This bind is supported by external conclusions \cite{brocki2022evaluation, dabkowski2017real}, and because evaluated properties are not the same (faithfulness vs human alignment), it demonstrates that human-independent metrics could replace a comparison with handcrafted ground-truth, thus reducing the workload of human annotators.
 
Overall, based on the saliency methods tested in the study, we conclude that XAI approaches cannot be universally ranked, and no method can be used without preliminary evaluation of its performance. This is further enforced by similar works using human-independent metrics: whether the articles conjointly proposed a new XAI method (and are possibly biased) or not, the published rankings of saliency-based methods are not only different from ours, but also between them \cite{anders2020fairwashing, fong2017interpretable, hooker2018evaluating, kapishnikov2019xrai}. In addition to the metric design itself and the application, the lack of consensus observed could also be explained by the variety of image modalities (natural vs. medical), or the inconsistent range of pixel values across all public or private datasets.

\section{Conclusion}

In this paper, we proposed an update on perturbation-based Fidelity metrics for evaluation of XAI methods, targeted at lowering the confidence of the network instead of its performance. This change follows the definition of a new constraint for few-class classification tasks, and enables use of the metric independently of the number of classes or amount of data, making it better adapted for real-world practical conditions. Our second contribution is a personalized generation and selection framework, ensuring that the applied perturbation is neither outside the dataset distribution or unable to provoke the uncertainty of the model. The framework is heuristic, transparent, and replaces fixed-value pixel replacement without further work, so it can be employed in future Fidelity metric changes. Measuring performance of the alteration candidates may be currently slow, but displayed a trade-off between generation speed and perturbation precision, that can be later optimized. We finally used the proposed approaches to rank well-known XAI methods on two few-class applications, and compared the hierarchies with human-driven object detection experiments. Global concordance of conclusions demonstrated that our Fidelity concept modifications, in addition to answering described obstacles, do not diminish the metric ability to compare the model reasoning with computed attributions. Finally, these results emphasized the lack of a generally-best XAI method, the need to conduct a comparison of methods for explaining any newly-trained CNN, and therefore the requirement 
of a fast and precise evaluation pipeline.

\printcredits



\bibliographystyle{cas-model2-names}
\bibliography{biblio}

\end{document}